\documentclass[10pt, a4paper]{article}
\usepackage{lrec2022} 
\usepackage{multibib}
\newcites{languageresource}{Language Resources}
\usepackage{graphicx}
\usepackage{tabularx}
\usepackage{soul}
\usepackage{titlesec}
\usepackage{draftwatermark}
\SetWatermarkText{(c) Mikhalkova et al. 13 July 2023\\Arxiv Preprint}
\SetWatermarkAngle{-90}
\SetWatermarkScale{0.2}
\SetWatermarkHorCenter{0.94\paperwidth}
\titleformat{\section}{\normalfont\large\bf\center}{\thesection.}{1em}{}
\titleformat{\section}{\normalfont\large\bfseries\center}{\thesection.}{1em}{}
\titleformat{\subsection}{\normalfont\SmallTitleFont\bfseries\raggedright}{\thesubsection.}{1em}{}
\titleformat{\subsubsection}{\normalfont\normalsize\bfseries\raggedright}{\thesubsubsection.}{1em}{}
\renewcommand\thesection{\arabic{section}}
\renewcommand\thesubsection{\thesection.\arabic{subsection}}
\renewcommand\thesubsubsection{\thesubsection.\arabic{subsubsection}}

\usepackage{epstopdf}
\usepackage[utf8]{inputenc}

\usepackage{hyperref}
\usepackage{xstring}

\usepackage{color}

\title{Psycho-linguistic Experiment on Universal Semantic Components of Verbal Humor: System Description and Annotation}

\name{Elena Mikhalkova, Nadezhda Ganzherli, Julia Murzina} 

\address{Tyumen State University \\
         625003, Volodarskogo, 6, Tyumen, Russia \\
         e.v.mikhalkova@utmn.ru \\}

\abstract{
Objective criteria for universal semantic components that distinguish a humorous utterance from a non-humorous one are presently under debate. In this article, we give an in-depth observation of our system of self-paced reading for annotation of humor, that collects readers’ annotations while they open a text word by word. The system registers keys that readers press to open the next word, choose a class (humorous versus non-humorous texts), change their choice. We also touch upon our psycho-linguistic experiment conducted with the system and the data collected during it.
\\ \newline \Keywords{trigger, self-paced reading, SPReadAH, linguistic annotation, humor, general theory of verbal humor, psycho-linguistic experiment, computational linguistics} }

\begin{document}

\maketitleabstract

\section{Introduction}

Universal semantic components of verbal humor that distinguish it from other genres and figures of speech have been widely discussed theoretically and analyzed by scholars based on documented examples of oral and written speech. Among these components are {\em the script opposition}, co-existence of two semantically opposing scripts in the same utterance, and {\em the trigger}, a switch between them. This condition of semantic clash is the foundation of the incongruity theory. For a long while, the idea of scripts and the trigger rested on experts' judgment about it.

Computational humor is a field of study that focuses on computer algorithms for humor detection and generation, the former being a more widely spread and considered to be a step preceding the latter. The data-driven approach to automatic detection of verbal humor presupposes that a large enough collection of humorous texts and an efficient algorithm will lead to computers understanding humor on par with human readers. The important difference between a human reader and an expert from academia is that, like an expert, the reader carries the skill of speaking the language and the cognitive mechanism of humor appreciation (sense of humor), but he or she does not have a theoretical paradigm in mind to read into a text while analyzing it. Annotation of a large enough collection by non-theorists is not a trivial task, not only because it requires a lot of time, annotators and a model for annotation that reflects the studied phenomenon. In regard to humor detection, to create such a model we would need theoretical conditions that are {\em sufficient} to differentiate between a humorous and non-humorous utterance.

One of the most known models, already implemented in automatic interpretation of humor, is the General Theory of Verbal Humor by V. Raskin and S. Attardo~\cite{raskin1987linguistic,salvatore1994linguistic}. The incongruity is one of its key components. However, according to other fundamental research, a semantic clash (incongruity) can be also found in tales, myths, narrative humor, metaphors, irony, pun, propaganda. In ``Éléments pour une théorie de l'interprétation du récit mythique'', A. J. Greimas~\cite{greimas1966elements} observed that myths and tales possess a similar feature: {\em renversement de la situation}, a reversal in a situation that splits the text into two planes:

{\em avant contenu inversé} before reversal

--- *not* ---

{\em après contenu posé} after reversal~\footnote{\url{https://www.persee.fr/doc/comm_0588-8018_1966_num_8_1_1114}}

Later V. Morin applied this paradigm to funny stories in ``L'histoire drôle''~\cite{morin1966histoire}. She enlisted some semantic oppositions like {\em Beau / idiot; Intelligent / laid; Papa /maman} Handsome / idiot; Intelligent / ugly; Dad / mom.

As for metaphors, \cite{richards1936philosophy} described the structure of a metaphor as a combination of a {\em tenor} (the target of metaphor), {\em vehicle} (the source, an image that resembles the tenor) and {\em ground} (what a tenor and vehicle share, their common features enabling metaphoricity). According to~\cite{zhu2017origin}, this structure has been known in the rhetoric since the Middle Ages as {\em primum comparationis}, {\em secundum comparationis}, {\em tertium comparationis}, where the first two are semantic scripts and the third one is something they share (verbalized, for example, with a polysemous word). \cite{lakoff2008metaphors} suggested a similar structure for conceptual metaphors: {\em target} and {\em source}. Irony is widely discussed as a speech genre where there is ``a proposition and an ironic statement that expresses the speaker's opposite attitude towards the proposition''~\cite{van2017can}. I.e. in irony the two scripts, {\em proposition} and {\em ironic statement} (literal and figurative), are opposed.

At the same time, \cite{roberts1994people} note that ``many rhetorical devices.. are not really non-literal'', i.e. it is not a figure of speech, or trope, that distinguishes figurative speech from literal. Rather, it is some words that in a certain context acquire a non-literal meaning. Figurativeness (metaphoricity) is also bound to familiarity: if a figurative word or expression becomes ubiquitous, a new meaning adds to the stock vocabulary~\cite{gentner2001convention} and does not require an explanatory context anymore. 

It follows that figurative speech occurs when a word or phrase is used in a non-literal original meaning due to a context that fosters it. Consequently, we deal with ``two planes of meaning''~\cite{krikmann2009similarity} simultaneously in one utterance. Let us consider three short texts that represent three different speech genres, but are all characterised by two planes of meaning.

\begin{itemize}
    \item[(1)] \textbf{Pun.} I could tell you a chemistry joke, but I know I wouldn't get a reaction.~\footnote{The example is from the data set for SemEval~\cite{miller2017semeval}.}
    \item[(2)]\textbf{Irony.} Oh hello flu! Thank you for fooling me in thinking you were gone!~\footnote{The example is from the data set for SemEval~\cite{van2018semeval}}
    \item[(3)]\textbf{Metaphor.} It's been a real circus at home since mom went on vacation.~\footnote{\url{https://quizizz.com/admin/quiz/5ace470abbf365001a6d3a0f/choose-the-figurative-language-example}}
\end{itemize}

The texts above contain words that can be understood in two meanings simultaneously: ``reaction'' in the pun means either chemical or emotional reaction, ``hello'' and ``thank you'' in the irony means ``sorry, you are here'', and ``circus'' in the metaphor means ``a mess''. The difference is that in puns the reader should understand the two meanings at once, but in irony and metaphor they should prefer the secret, non-literal meaning. Nevertheless, the three texts possess enough context for some words to be treated as ambiguous. Hence, we can speak of two topics stretching through the text and endorsing ambiguity. Figurativeness -- the property of the text being suggestive of two interpretations -- is gradually accumulated in such texts.

Hence, it is quite clear that humorous texts differ from texts with no incongruity in them, but it is not clear how they differ from texts with semantic ambiguity like metaphors and irony. In~\cite{mikhalkova2019comparison}, we tested computer algorithms designed to detect one type of figurative speech (either metaphor, or irony, or puns) on other types of figurative speech and found out that they can be equally effective, i.e. they do not see much difference between these types.

In this paper, we propose a system of self-paced reading (SPReadAH: Self-Paced Reading for Annotation of Humor) when the reader opens a text word by word and marks the place where he or she starts to think of the text as humorous. We also briefly describe an experiment on manual annotation of the trigger by Tyumen State University students with no research and terminological background in the incongruity theory of humor. We conclude about perspectives of the practical search for universal semantic components of humor.

\section{Related Work}

As for experimental research on metaphoricity perception, there are a lot of studies with application of sensors on how the human brain reacts to ambiguous texts~\cite{coulson1998frame,macgregor2020neural,bekinschtein2011clowns}. In them, the term ``semantic ambiguity'' often refers to texts with metaphors that pose a difficulty for a reader as compared to more direct texts. \newcite{coulson1998frame} showed experimentally that texts with frame-shifting take longer to be read than texts with only one frame (script). \newcite{bambini2012metaphor} gave an extended account of further neurological research that mainly focuses on different mechanisms that are activated in the brain when it processes metaphors. They called this branch of research experimental and cognitive pragmatics.

There is a lot of experimental research on texts that make readers laugh or smile. There is evidence of a different reaction to texts with two scripts: they take longer to read, invoke various mimics, change direction of gaze. \newcite{vaid2003getting} observed an experiment establishing the time course of two scripts activation: the initial and the intended meanings. The latter is activated by the joke and is semantically incongruent, opposing the original meaning in verbal humor. In a later work, \newcite{coulson2006looking} studied eye movements in the course of reading two types of sentences: humorous with a punch line at the end of a sentence, and non-funny texts in which the funny word was changed so that the humor was lost. The experiment showed that the overall time of eye fixation on a humorous text is longer than that on non-funny texts. In a recent study~\cite{gironzetti2016smiling} and a report ``Mutual Gaze and Gaze Aversion in the Negotiation of Conversational Humor'' (delivered at the conference of the International Society of Humor Studies, 2019), E. Gironzetti described the mimics (smile) and gaze (eye fixation, and its direction) connected with humor in a dialogue. \newcite{lopez2017psycholinguistic} pinpoint the deficiencies of studying humor with self-paced reading methods as compared to eye-tracking. Being unable to move to the previous word is listed as one of the main drawbacks of self-paced reading. Combining it with eye-tracking helps to establish at what moment the reader figures out that he or she is reading a humorous text.

\section{System Description and Annotation Process}

Presently, there is a gap between theoretical understanding of humorous incongruity and experimental proof with a general audience that it even exists in contrast not only to direct speech, but to other ambiguous, metaphorical texts. We want to experimentally observe what happens in the process of text comprehension when readers realize they deal with humor. For this we need to register turning points in their reading experience, precisely the moment of realization and to what class (humor or not) they think the text belongs. Also, we want to register if readers change their mind, because they can be misled by allusions, their reading background, etc. For this purpose we have designed a specialized program with self-paced reading, i.e. when ``a reader presses a button to see each successive word in a text''~\cite{just1982paradigms}. SPReadAH (Self-Paced READing for Annotation of Humor) is built using the Python programming language powered by PsychoPy library~\cite{peirce2007psychopy,peirce2019psychopy2}. It includes the following main blocks:

\begin{enumerate}
    \item Instructions for respondents.
    \item A dummy annotation test to check whether the respondent understands the task and which keys to press for which decision.
    \item Annotation of short texts.
    \item Evaluation of funniness of the annotated texts.
\end{enumerate}

Instruction includes a short explanation of the experiment's purpose, plain definitions of classes to be labelled and mechanics of labelling texts (which keys to press). 

The dummy annotation test includes annotation of four sample texts (one text of each class). A respondent tries labelling texts and gets used to navigation and screen prompts.

\begin{figure}[!h]
\begin{center}
\includegraphics[scale=0.28]{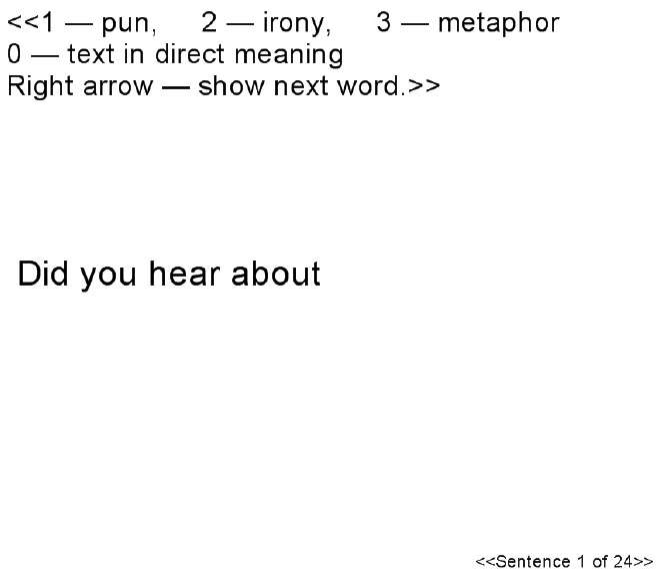}
\caption{Navigation to the next word and class assignment.}
\label{fig:eval}
\end{center}
\end{figure}

The next stage is the experiment proper when respondents open texts word by word (meaning written words: any symbols between two spaces) by pressing a key. Before start, the texts for annotation are shuffled randomly. The respondent navigates to the next text and, while opening it, he or she should decide whether the text belongs to one of the classes (humorous versus non-humorous). Different keys are used for navigation to the next word (right arrow) and the next text (space). If the word was opened, it stays on the screen until navigation to the next text, see Fig.~\ref{fig:eval}. Here and further in figures, the example annotation is given in English. The example is taken from the data set for SemEval~\cite{miller2017semeval}.

In this respect, our experiment is different from the classic self-paced reading experiments showing just one word at a time. Also, the respondent can only make decisions about the last word opened, i.e. they cannot go back and annotate words and texts before it. During the experiment it appeared that sometimes respondents were in a hurry to be over with the experiment. To force them to think more about the text we included a one second delay before respondents can proceed to the next word.

\begin{figure}[!h]
\begin{center}
\includegraphics[scale=0.2]{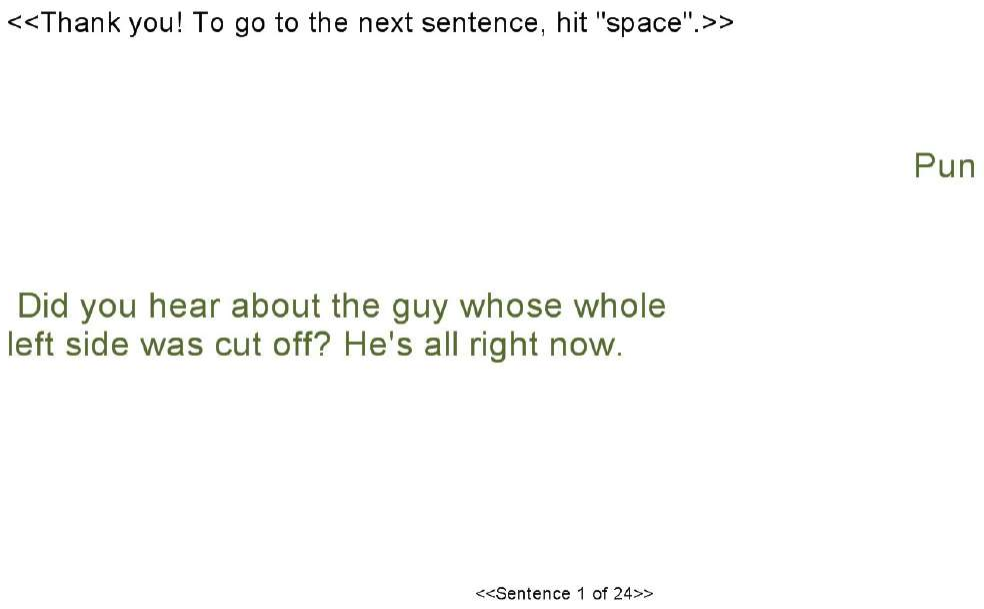}
\caption{When the text is fully shown and class chosen.}
\label{fig:answered}
\end{center}
\end{figure}

If respondents start to believe that the text they are reading belongs to one of the classes, they press a corresponding key, and the relevant label is shown on the screen. This decision may be changed by pressing another relevant key at any next word but before going to the next text, see Fig.~\ref{fig:answered}.

\begin{figure}[!h]
\begin{center}
\includegraphics[scale=0.28]{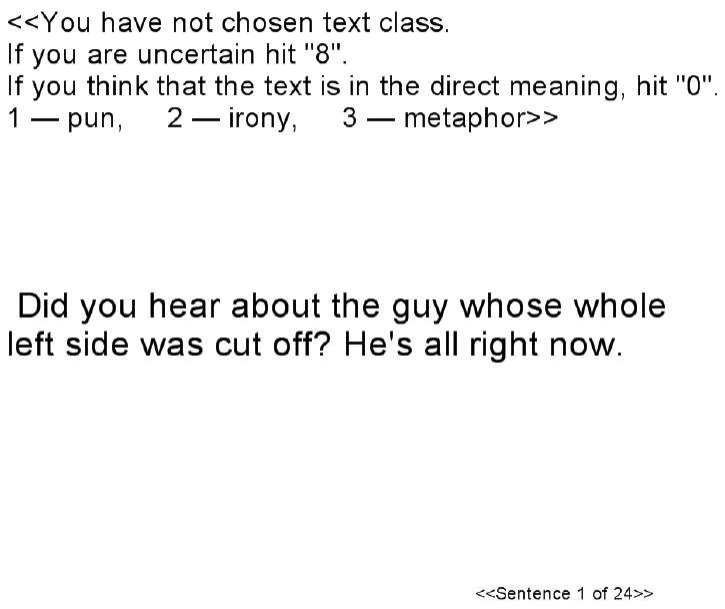}
\caption{Option for texts with no class chosen.}
\label{fig:noclass}
\end{center}
\end{figure}

When the whole text is on the screen, respondents have an opportunity to change or finally confirm their decision. In case no class is chosen, the system asks to confirm that respondents cannot attribute this text to any class. Otherwise, they cannot proceed to the next text, see Fig~\ref{fig:noclass}.

\begin{figure}[!h]
\begin{center}
\includegraphics[scale=0.2]{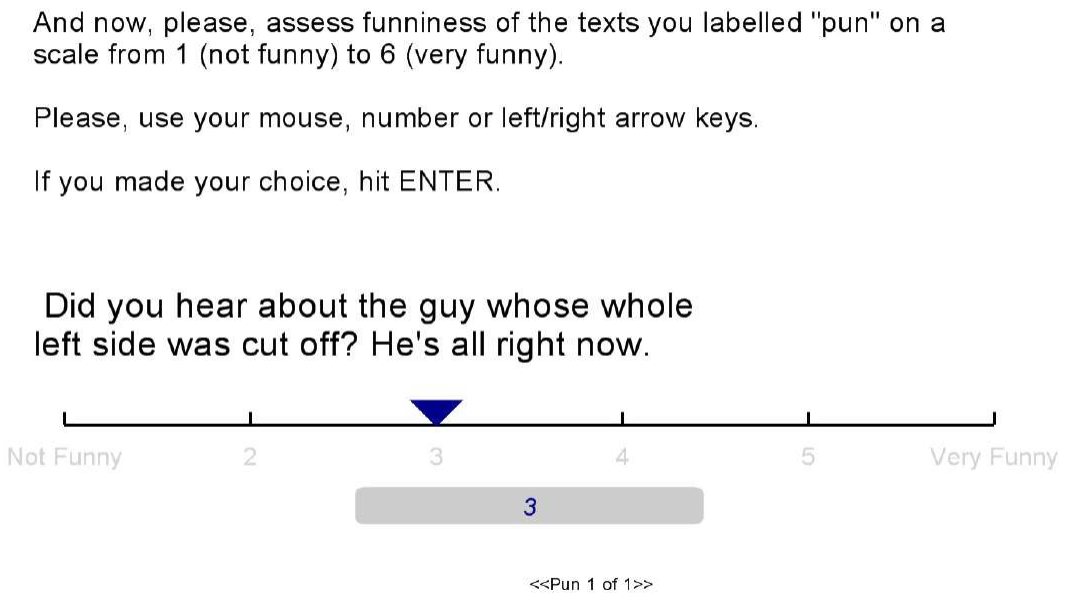}
\caption{Evaluation of ``funniness''.}
\label{fig:funniness}
\end{center}
\end{figure}

The last part of the experiment is evaluating funniness of the texts that were attributed to a humorous genre (class). The funniness is assessed on a scale from 1 (not funny) to 6 (very funny); the even score makes sure that respondents do not pick up a ``lukewarm'' middle score in ambiguous cases. At this stage, one can use number keys, the mouse, or arrow keys, see Fig.~\ref{fig:funniness}.

In the course of experiment, SPReadAH logs all the events in a separate .log file to be later processed by researchers. The log files are available at the project page: \url{https://github.com/Na-Gan/SPReadAH_Project}.

\section{Experiment Setup}

To mark a trigger, i.e. find a place in a text that persuades the reader this text is humorous, is such a task in which it is important to check that readers do not see triggers in, first, texts of completely non-humorous nature and, second, texts that resemble humor, e.g. metaphorical speech, irony, etc. For our linguistic experiment with SPReadAH we chose texts of four classes: metaphor, irony, a pun, none of the above. Puns (jokes based on wordplay) are the only humorous genre in the set. The texts were mainly collected from the Russian literature: ``The Adventures of Dennis'' by V. Dragunsky, ``Seventeen Moments in Spring'' by Yu. Semyonov, ``Pelageya and the White Bulldog'' by B. Akunin. We also included Russian tweets with hashtag \#irony (the hashtag was removed in the experiment) and Russian puns from several Web resources. The texts will be gradually added at the project page.

Our linguistic experiment was conducted on-line via TeamViewer software in July and August, 2020. 27 students from different University departments took part in it. It included five series of 24 short texts: 4 texts with puns, irony, and metaphors and 12 texts without any of them (the {\em None}-class) -- 120 short texts in total. A respondent annotated only one series per day. Each time before annotation they filled out a small anonymous on-line Google form stating their background information (sex, age, education, native language, etc.) and their current well-being (attitude to the coming work, mood, etc.). The survey can show a correlation between the mood and disagreement of certain respondents who, for example, were unwilling to read texts attentively and annotated them randomly or annotated the first word every time.

The results of the 2020 experiment are still under survey. However, our earlier work without SPReadAH (2018-2019) showed several peculiarities in the annotation. First, nearly 50 \% of texts in direct meaning (the {\em None}-class) were assigned to some other non-direct speech types. It may be that the texts were not purely direct, and we overlooked it. Or, due to polysemy, it is easy to see an indirect meaning in any text. Second, the respondents agreement about the trigger word was always quite low (Fleiss' kappa in one series was 0.453 maximum). However, if we grouped together several words near the most tagged word, the agreement would rise significantly. This poses a question whether the trigger is one word or it is a group where the figurativeness (metaphoricity) accumulates so much that the text becomes indirect. For annotators with a different background this ``triggering'' moment can occur on one of the words from the group.

\section{Conclusion}

Our experiment on the annotation of the trigger word in short humorous texts aims at filling the gap between the theoretical assumptions about its existence and their experimental proof. In the current research, we have described the system of self-paced reading for the annotation of humor: SPReadAH. The system is designed for the particular task of classifying between humorous and non-humorous texts and stores information about readers' choice, thus, marking a place in the text where they decide they deal with humor. We have also briefly described our experiment conducted in summer 2020, as its results are still under survey. We plan to, first, design methodology of calculating the position of the trigger word as it can be not just a word, but a word group, and, second, to define whether our respondents do see difference in incongruity of humorous texts as opposed to any other figurative speech. A more distant plan includes eye-tracking that facilitates finding the trigger.
\section{Bibliographical References}\label{reference}

\bibliographystyle{lrec2022-bib}
\bibliography{lrec2022-example}


\end{document}